\documentclass[letterpaper, 10 pt, conference]{ieeeconf}  

\IEEEoverridecommandlockouts                              
\makeatletter
\def\bstctlcite{\@ifnextchar[{\@bstctlcite}{\@bstctlcite[@auxout]}}
\def\@bstctlcite[#1]#2{\@bsphack
  \@for\@citeb:=#2\do{%
    \edef\@citeb{\expandafter\@firstofone\@citeb}%
    \if@filesw\immediate\write\csname #1\endcsname{\string\citation{\@citeb}}\fi}%
  \@esphack}
\makeatother

\usepackage[vlined, ruled, linesnumbered, commentsnumbered]{algorithm2e}
\usepackage{amsmath}
\usepackage{cite}
\usepackage{color}
\usepackage{xcolor}
\definecolor{chred}{rgb}{0.8,0,0}
\definecolor{chgrey}{rgb}{0.5,0.5,0.5}
\usepackage{graphicx}
\usepackage{caption}
\usepackage{subcaption}
\usepackage{dblfloatfix}
\usepackage{eqlist}
\usepackage{txfonts}
\usepackage{url}
\usepackage{footmisc}
\usepackage{booktabs}
\usepackage{balance}\usepackage{flushend}

\usepackage{multirow}
\usepackage[para,online,flushleft]{threeparttable}
\title{\LARGE \bf Designing a Mechanical Tool for Robots with 2-Finger Parallel Grippers}

\author{Zhengtao Hu$^{1}$, Weiwei Wan$^{1,2,*}$, and Kensuke Harada$^{1,2}$
\thanks{$^{1}${Graduate School of Engineering Science, Osaka University, Japan.}
$^{2}${National Inst. of AIST, Japan.} *{Correspondent author: Weiwei Wan,}
{\tt\small wan@sys.es.osaka-u.ac.jp}}%
}

\begin{document}
\maketitle
\thispagestyle{empty}
\pagestyle{empty}

\begin{abstract}
This work designs a mechanical tool for robots with 2-finger parallel grippers, 
which extends the function of the robotic gripper without additional requirements 
on tool exchangers or other actuators. The fundamental kinematic structure of the 
mechanical tool is two symmetric parallelograms which transmit the motion of 
the robotic gripper to the mechanical tool. Four torsion springs are attached 
to the four inner joints of the two parallelograms to open the tool as the 
robotic gripper releases. The forces and transmission are analyzed in detail 
to make sure the tool reacts well with respect to the gripping forces and 
the spring stiffness. Also, based on the kinematic structure, variety 
tooltips were designed for the mechanical tool to perform various tasks. 
The kinematic structure can be a platform to apply various skillful gripper 
designs. The designed tool could be treated as a normal object and be picked up 
and used by automatically planned grasps. A robot may locate the tool through 
the AR markers attached to the tool body, grasp the tool by selecting an 
automatically planned grasp, and move the tool from any arbitrary pose 
to a specific pose to grip objects. The robot may also determine the optimal 
grasps and usage according to the requirements of given tasks.
\end{abstract}

\section{Introduction}

Manufacturing requires fast reconfiguration of robotic systems to adapt to various products. 
Especially in the process of assembly, it is a challenge for robots to process a variety of 
components fast and precisely. Thus, developing robotic systems to handle a wide range of objects
in a reliable and low-cost way is highly demanded. In the past decades, various robotic hands with 
advanced functions were proposed. Each of them had merits in certain aspects. Also, 
to deal with different objects, tool changers and finger-tip changers were designed 
to expand the feasible grasp scope.

While the tool changers and finger-tip changers increased the flexibility of robot systems,
their drawbacks are also obvious. They require power supply, vacuum supply, or delicate mechanism and control 
to assure the firm connection between the actuators and a robot end. The tools and finger-tips have to be
designed especially for specific robots and switchers.

\begin{figure}[!htbp]
    \centering
    \includegraphics[width=.9\columnwidth]{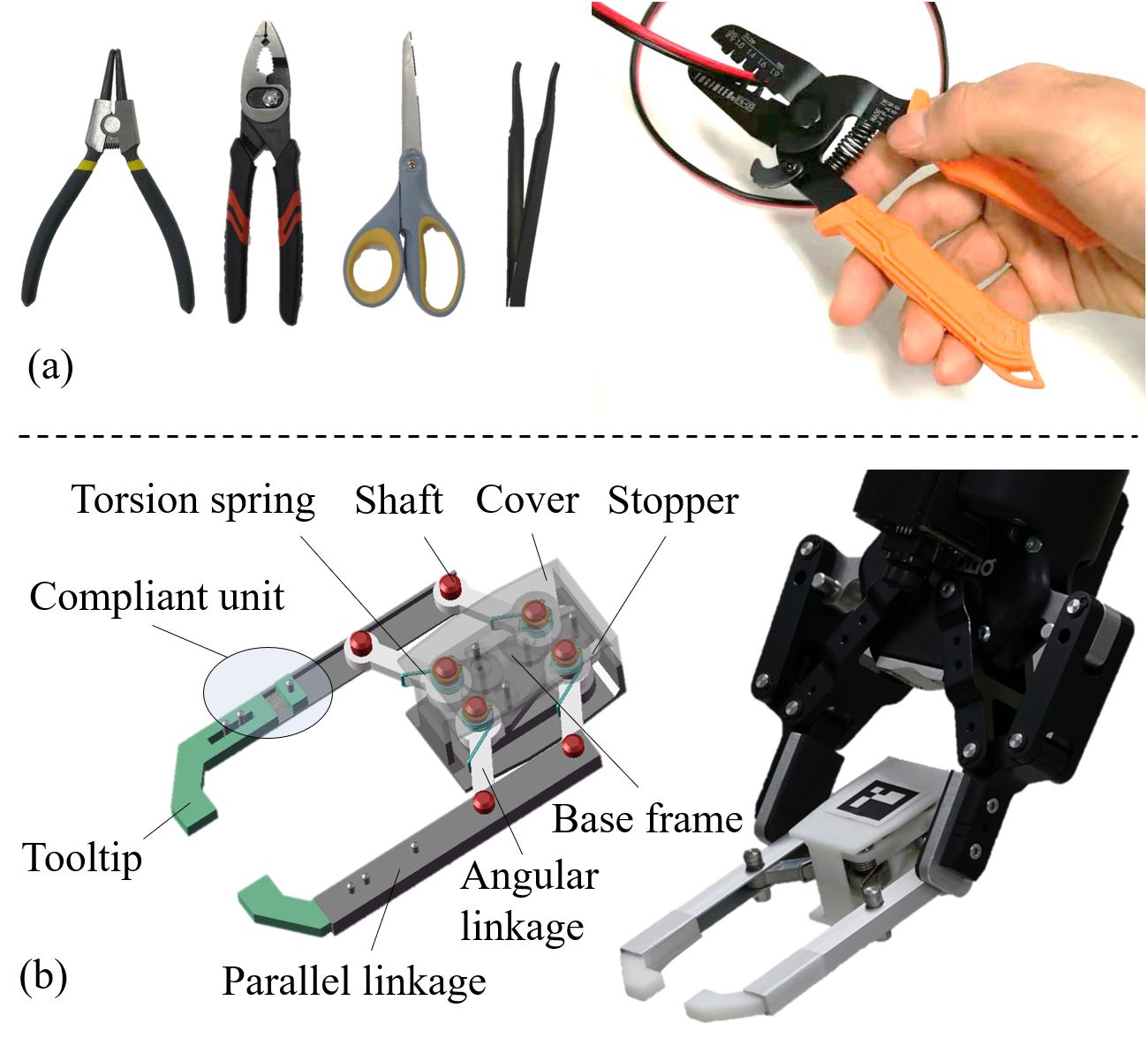}
    \caption{(a) Various tools designed for human hands.
    (b) A mechanical tool designed for parallel robotic grippers. 
    The tool is purely mechanical. There are no additional requirements
    for power cables or air tubes. Any robots with parallel grippers could use it.}
    \label{teaser}
\end{figure}

In this paper, we propose a solution by designing a mechanical tool for robots with 2-finger parallel grippers (Fig.\ref{teaser}(b)).
Like the many tools designed for human hands (Fig.\ref{teaser}(a)), the mechanical tool is general and independent from
specific robots. Any robot with 2-finger parallel grippers could recognize, grasp, and use the tool.
The tool could have lots of variations in the tooltips. A robot may select and use different ones 
to finish different tasks. The tool is purely mechanical. There are no additional requirements
for power cables or air tubes. There is also no special requirements for robotic end-effectors. 
The tool could be used by any robots with 2-finger parallel grippers.

The features of the design are: 1) The tool is mechanical and is only manipulated and
actuated by robotic grippers. 2) The tool can be designed with various tooltips 
adapted for different tasks. 3) The tool can be placed at an arbitrary pose
in the work space, and be recognized, grasped, manipulated, and used by
parallel robotic grippers. 

In the following sections, we will discuss the details of the design, including
the kinematic structure, the analysis and optimization of grabbing force and sizes,
and the consideration of stable placements, recognition, pose adjustment, and working poses.
We carry out experiments to analyze the performance of the design, as well as develop
a robot system that uses the tools with different tips to pick up various objects. 
The experiments and analysis show that the mechanical tool is a flexible alternative to 
tool changers and finger-tip changers. With the help of visual detection and motion planning 
algorithms, robots are able to automatically recognize and use the tool to finish a wide range of tasks.

\section{Related Work}

We in this section review the related studies by separating them
into three categories: 
(1) The design of robot hand changers, (2) The design of versatile and adaptive grippers, and 
(3) Grasp and regrasp planning. 

\subsection{The design of robot hand changers}
Robot hand changers originate from the tool changers used in 
Computer Numerical Control (CNC) machines \cite{gokler1997design}\cite{lundberg2011automatic}\cite{rogelio2014development},
and are still widely studied \cite{ryuh2006automatic}\cite{kordi2007development}.
The reason is to use robots in industry applications, engineers have to design various
grippers \cite{monkman2007robot} to adapt to different tasks and objects.
 
Recent development in robot hand changers has two trends. The first is developing
automatic tool changers for mobile manipulators. Some of them are electro-mechanically actuated,
like the one presented in \cite{gyimothy2011experimental}. 
Some others are passive, like the one presented in \cite{berenstein2018open}.
Which used passive mechanisms actuated by
host robots. Other than the changers, some studies design interfaces for robot end-effectors.
An example is the da Vinci Surgical Research Kit \cite{mckinley2016interchangeable}. The idea is to set an adapter
between the tool and the original end-effector. A finger-tip changer \cite{clevy2005micromanipulation} shares the similar idea.

The aforementioned studies all provide effective ways to change hands for robots. 
However, though those systems can provide reliable and precise fixing as 
well as connection, the efficiency of the exchanging process is still a 
problem. Also, the tools are adapted for specific end shapes,
and the peripheral equipment is indispensable, which 
restricts the potential applications. Unlike them,
we in this paper design a mechanical tool for parallel robot grippers.
The design is passive and does not need any power or air supply.
It is general and could be grasped and used by any robots with parallel grippers. 
To our best knowledge, this is the first work that designs a general mechanical tool for robots.

\subsection{The design of versatile and adaptive gripper}
Other than hand changers, more studied are devoted to designing versatile and adaptive gripper.
For example,  Harada et al. \cite{harada2016proposal} developed a novel gripper by
combing a multi-finger mechanism and a granular jamming component, which can 
achieve versatile grasping firmly as well as flexibly. 
Triyonoputro et al.\cite{triyonoputro2018double} developed a double jaw 
hand for grasping and assembly. The inner and outer grippers can work together 
to finish a task. The design was inspired by a human hand holding and 
manipulating two objects using one hand in product assembly. 
Nie et al. \cite{nie2019hand} proposed a pair of fingers for arranging 
screws. The gripper can pick up and tile a screw to let the screw slide to the 
bottom of the finger so as to achieve the picking and alignment. 
Hirata et al. \cite{hirata2011design} proposed a gripper with tips 
that can cage and self-align objects. 
Laliberte et al. \cite{laliberte2002underactuation} developed an under-actuated 
robot hand driven by two motors. The fingers of the hand could adapt to
the shape of objects during grasping. The hand is essentially a gripper but is 
versatile and adaptive.
Ma et al. \cite{ma2016m} presented a gripper with one underactuated finger and 
a passive finger, which is helpful to implement adaptive, shape-enveloping 
grasps, and also in-hand manipulation.

The versatile and adaptive grippers can perform well for a group of workpieces, 
however they can never be completely universal. Also, the structure has to be
carefully designed to expand the function. Compared to them, using parallel gripper
but preparing a range
of mechanical tools could be a better way to obtain versatility and adaptivity.

\subsection{Grasp and regrasp planning}

Besides the design, this work also uses automatic recognition and motion planning to
recognize the tool, and plan a motion to adjust and use the tool.
The automatic recognition and motion planning
is based on our previous studies in grasp and regrasp planning.
Some representative ones are as follows.
Wan et al. \cite{wan2017teaching} presented a method for motion planning to 
achieve smart assembly tasks. In the work, a 3D vision system was employed to detect 
human operation, point clouds and geometric constraints were used to find 
rough poses. A method was used to plan the motion of robots to reorient objects.
Raessa et al. \cite{raessa2019teaching} proposed a method to teach a dual-arm
robot to use common electric tools. Grasp and regrasp planning were implemented 
to adjust the work pose of the tool following \cite{wan2019preparatory}. Sanchez et al. 
\cite{sanchez2018arm} developed a planner with orientation constraints to manipulate 
a tethered tool. 
Beyond our group, similar studies about the grasp and regrasp planning could be found in
\cite{Dogar15}, \cite{Lertkultanon17}, \cite{xian2017closed}, \cite{suarez2018can}, \cite{chen2018manipulation}, etc.


\section{Design and Optimization}

This section presents the details of design and optimization, including the kinematic structure,
the analysis and optimization of forces and sizes, as well as the variation in tooltips.

\subsection{The kinematic structure}
\begin{figure}[htbp]
\begin{center}
\includegraphics[width=.9\columnwidth]{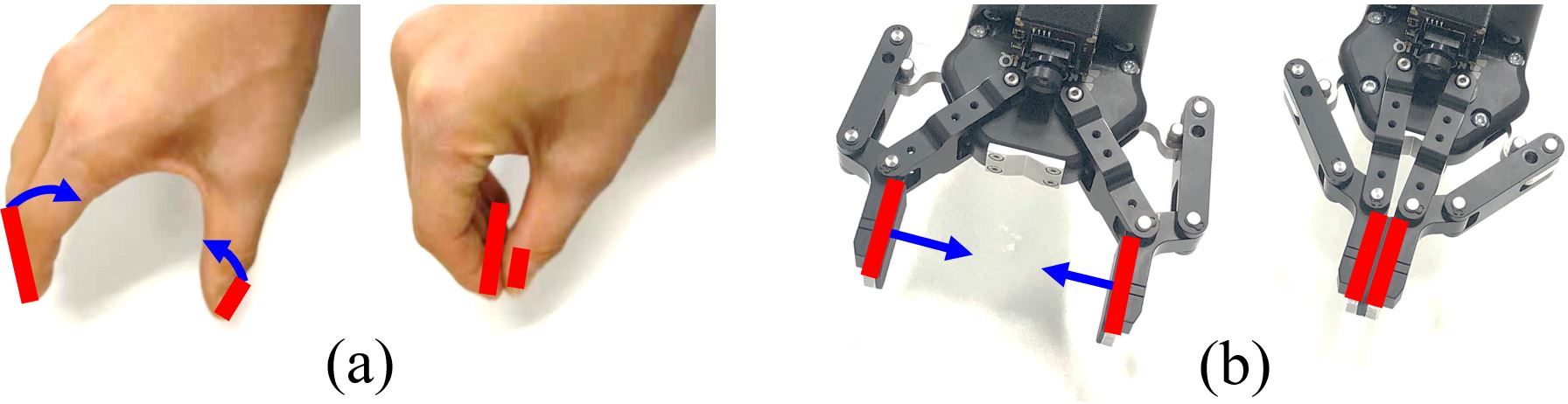}
\caption{(a) The main synergy of a human hand. The thumb and the remaining fingers form a rotational grab.
The tools designed for human hands usually have a rotational joint. (b) The motion of a parallel robotic gripper.
The tool designed for it is best to have a parallel mechanism.}
\label{humantorobothand}
\end{center}
\end{figure}

The tools designed for human hands usually have a rotational joint, 
as is shown in Fig.\ref{teaser}(a).
The reason is because the rotational grab formed by the thumb is the 
main synergy of human hands\cite{santello1998postural}, as is shown in Fig.\ref{humantorobothand}(a). 
Likewise, a tool designed for parallel robotic grippers (Fig.\ref{humantorobothand}(b))
is best to have a parallel mechanism
to cater the parallel motion of the robotic gripper.

\begin{figure}[htbp]
\begin{center}
\includegraphics[width=.95\columnwidth]{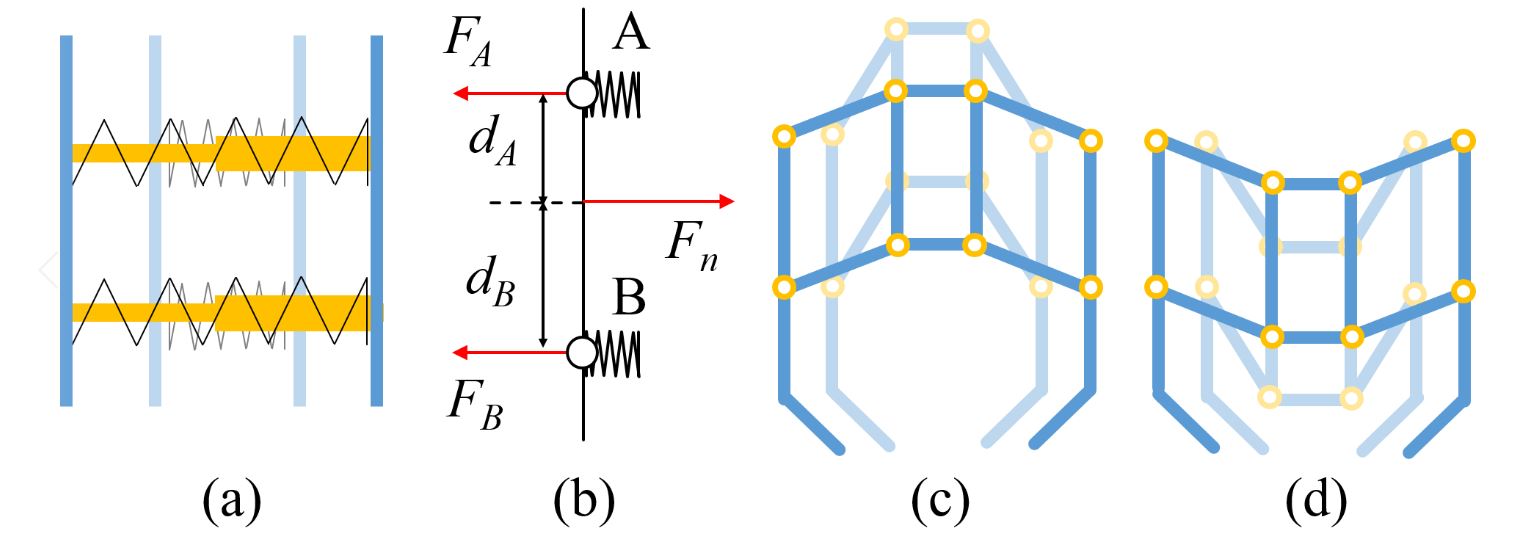}
\caption{(a) The motion of an intuitive parallel mechanism made by sliding rails and linear springs.
(b) The free body diagram of the intuitive mechanism. (c) A parallel mechanism made of two
symmetric parallelograms. In this case, the base frame will move backward while the tool 
is closed. (d) A similar parallel mechanism as (c). The difference is in this case, 
the base frame will move forward while the tool is closed.}
\label{parallel}
\end{center}
\end{figure}

An intuitive idea to implement parallel motion is to use sliding rails. Linear springs
may be attached to the rails to help return to the initial state after releasing.
Fig.\ref{parallel}(a) illustrates the intuitive mechanism.
This idea is easy to understand, but is difficult to assure stable parallel motion.
Fig.\ref{parallel}(b) shows the free body diagram of the intuitive mechanism. 
Since the sliding rail cannot bear force along the moving direction, 
$F_{A}$ and $F_{B}$ are all from the springs. While the fingers keep parallel, 
the two springs deform equally and $F_{A}$ is the same as $F_{B}$. To meet the momentum equilibrium equation 
$F_{A}d_{A}-F_{B}d_{B}=0$, $d_{A}$ and $d_{B}$ must equal. That means to assure a stable parallel motion,
the contact can only be applied at the center of the two springs, which severely decreases the 
possible grasp configurations and increases the difficulty of automatic
manipulation planning. 

Instead of the simple sliding rails, we design the tool by using two symmetric parallelograms,
as is shown in Fig.\ref{parallel}(c) and Fig.\ref{parallel}(d). 
The two parallelograms allow the force from robotic grippers to be evenly distributed to the joints,
and are therefore able to assure stable parallel motion.
Both of the two configurations in Fig.\ref{parallel}(c) and Fig.\ref{parallel}(d) can provide parallel 
motion transmission. However, for the scheme shown in Fig.\ref{parallel}(d), 
the grasp is likely to be impaired since the grasping space will be occupied by the 
forward moving base. Also, as the heaviest part of the tool, the forward motion of 
the base also impair the stability of grasping. Thus, the configuration in Fig.\ref{parallel}(c) 
is selected as the kinematic structure of our design.
Besides space, the configuration has another advantage, 
which will be explained in detail in the force analysis section.

\begin{figure}[htbp]
\begin{center}
\includegraphics[width=.9\columnwidth]{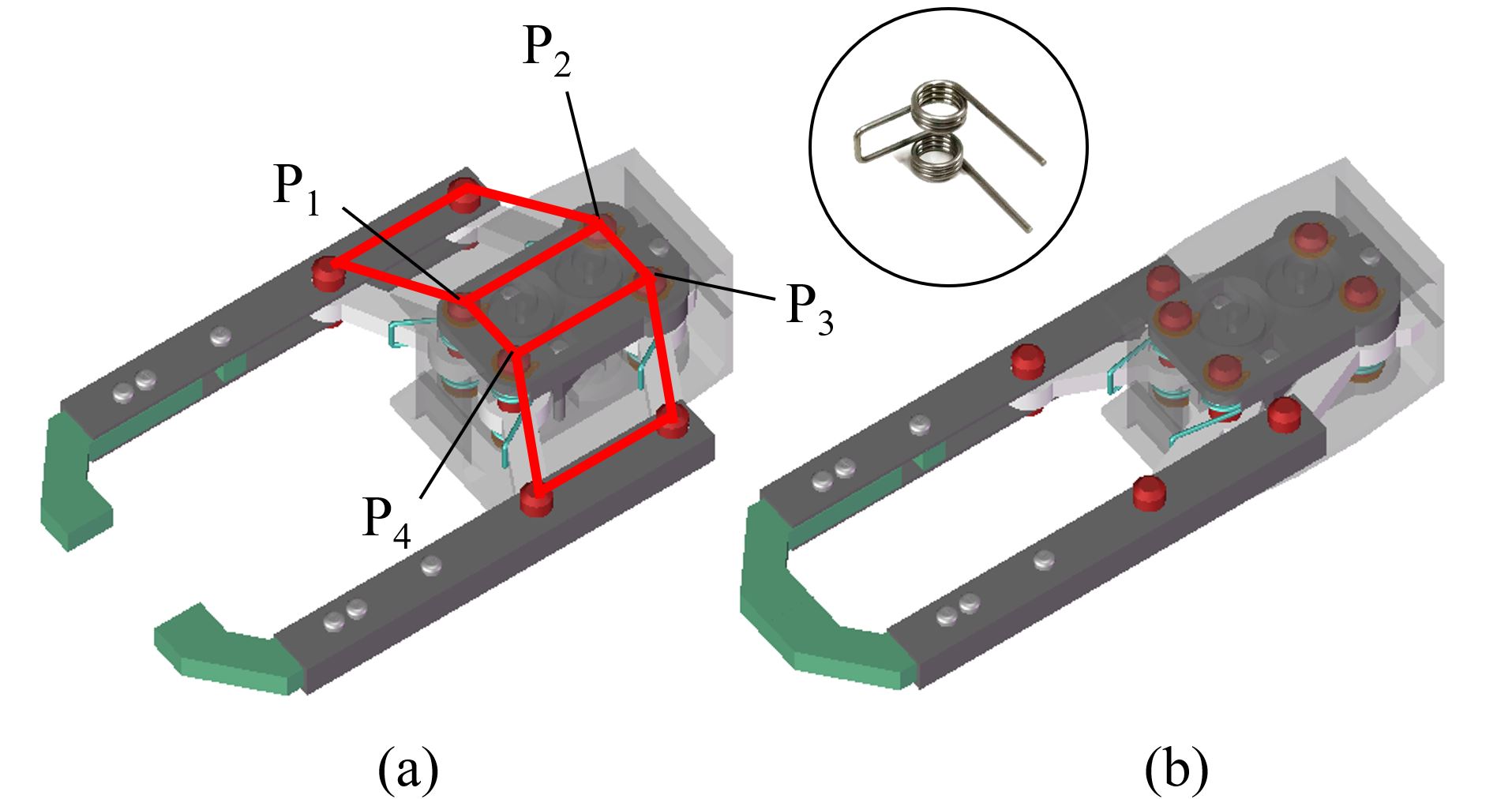}
\caption{The designed mechanical tool. (a) The tool is completely open. (b) The tool is closed.
Torsion springs shown in the circle are installed at joints P$_1$$\sim$P$_4$.}
\label{tool_two}
\end{center}
\end{figure}

Fig.\ref{tool_two} shows the design. In Fig.\ref{tool_two}(a), the jaw
of the tool is opened to its maximum. In Fig.\ref{tool_two}(b), the jaw is fully closed.
The two parallelograms are installed symmetrically. They force two tooltips to move in parallel. 
As is shown in in Fig.\ref{tool_two}(a), four torsion springs are installed at joints P$_1$$\sim$P$_4$. 
The torsion springs are concentric with their supporting shafts. 
The ends of the torsion springs are respectively fixed to the base frame and the angular linkages. 
The torsion springs provide friction to prevent the tool 
from dropping as the robotic gripper holds it. They also provide forces to open
the tool as the robotic gripper releases. 

The torsion springs are installed with a pre-angle $\beta$, which is determined by
the stopper crafted in the base frame. The force exerted by a spring to an angular linkage is therefore:
\begin{equation}
F_{spring}=\kappa(\beta +\Delta \theta) \label{eq-4}
\end{equation}
where $F_{spring}$ is the exerted force. $\beta$ is the pre-angle. $\kappa$ is
the elastic coefficient. $\Delta \theta$ is the rotational angle of the angular linkage.
Choosing a proper $\beta$ is an optimization problem. On the one hand,
with the same $\Delta \theta$, a large $\beta$ provides a large resistance force to
robot grippers and hence provides larger friction to prevent the tool from 
dropping out of the robot gripper. It also leads to a shorter stroke of the robotic gripper to
get the same transmitted force.
On the other hand, if $\beta$ is too large, the robot gripper has to exert a very large force to
overcome the tension of the torsion springs. In the worst case, the tool may not be closed.
The details of force analysis will be discussed in the next subsection.

\subsection{Force analysis}
In this subsection, we analyze the forces between the tool and 
a robot gripper to optimize the design.
The subsection comprises two parts. In the first part, we 
analyze the condition for a robot gripper to firmly hold the tool as well as
the relationship between robot grasping force and the resistance force from the torsion springs
when a robot gripper is holding the tool. In the second part, we analyze the maximum weight of
objects that can be pick up by the tool.

\subsubsection{Holding the tool}
We model the contact between the robot gripper and the tool as a soft contact.
Following \cite{howe1996practical}\cite{ciocarlie2007soft}, the force and friction 
exerted by the robot gripper can be computed by:
\begin{equation}
f^{2}+\frac{T^{2}}{e^{2}}\leqslant \mu^{2}F_{n}^{2}\label{eq-fa-1}
\end{equation}
where $f$ is the tangential force at the contact. 
$T$ is the torque at the contact. $F_{n}$ is the gripping force exerted by the robot gripper. 
$e$ is an eccentricity parameter computed by the ratio 
between the maximum friction and friction torque on the contact surface:
\begin{equation}
e=\frac{\max T}{\max f}\label{eq-fa-2}
\end{equation}

\begin{figure}[!htbp]
\begin{center}
\includegraphics[width=.9\columnwidth]{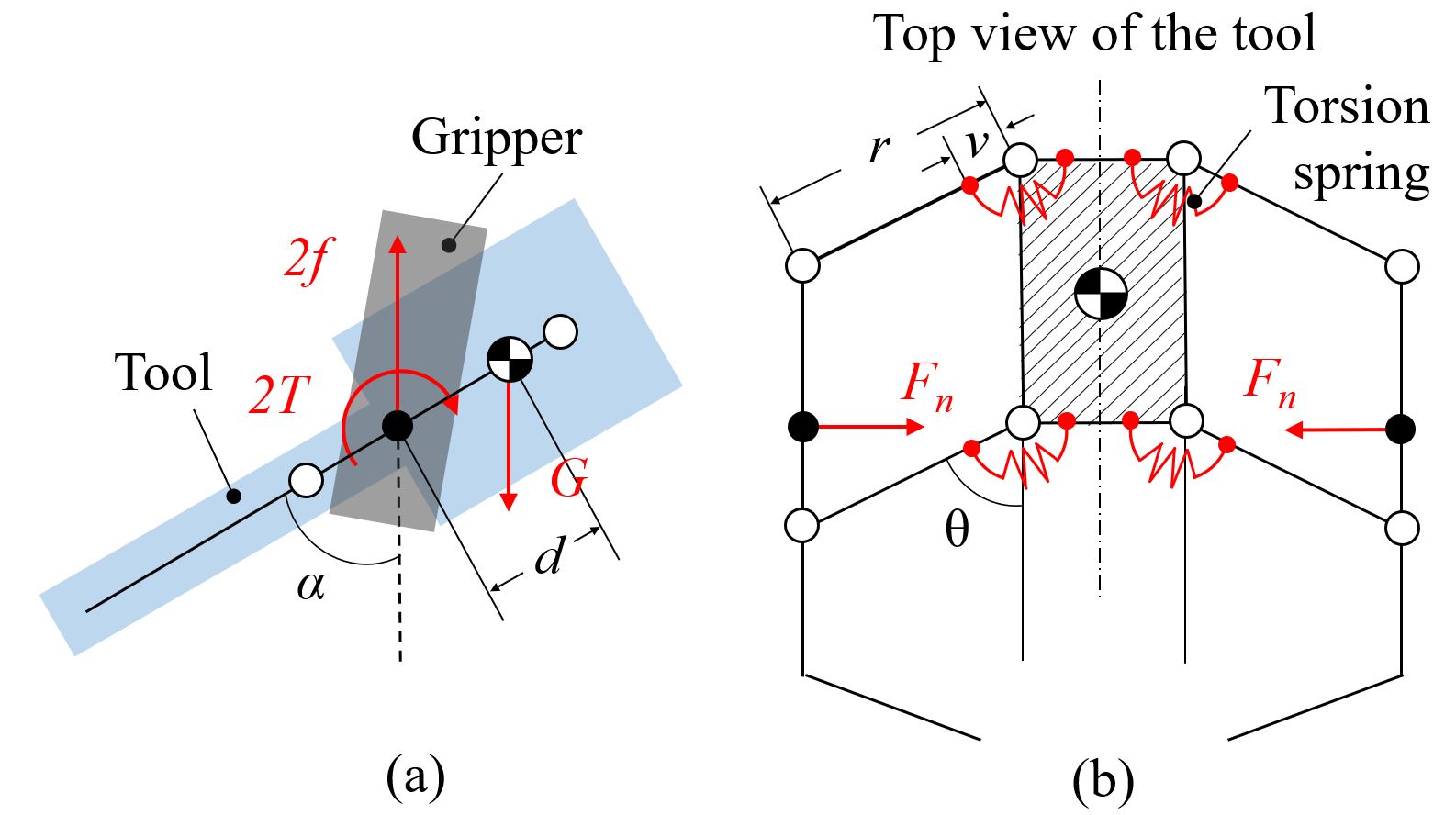}
\caption{The free body diagram when the tool is held by a robot gripper. 
$F_{n}$ is the force exerted by the robot gripper.}
\label{fbdheld}
\end{center}
\end{figure}

The free body diagram when the tool is held by a robot gripper is shown in 
Fig.\ref{fbdheld}. Here, $f$ is the friction force at the contact between the robot finger and the tool.
$T$ is the torque at the contact. $\alpha$ is the angle between the tool and the direction of gravity.
It is called the tool angle.
$d$ is the distance between the grasping point and the center of mass $com$ of the tool.
By using the symbols shown in the figure and the soft finger contact model,
we can get the condition to hold the tool as:
\begin{equation}
d\leqslant\max T\sqrt{\frac{4\mu^{2}F_{n}^{2}-G^{2}}{G^{2} \sin{\alpha}^{2}\mu^{2}F_{n}^{2}}} \label{eq-d-1}
\end{equation}
When $d$ equals 0, there is no torque at the contact. The robot gripper can hold
the tool as long as $2\mu F_n \geq G$. When $d$ is not 0, the $F_n$ needed to hold the tool
is a function of $d$, $G$, $\mu$, and $\alpha$.

If the tool could be held firmly, namely the friction and the friction torque on the surface between
the gripper and the tool are enough to prevent the tool from dropping,
the relationship between $F_{n}$ and the force
exerted by the torsional springs $F_{spring}$ is:
\begin{equation}
F_n=\frac{G\cos \alpha \tan \theta }{2}+\frac{2vF_{spring}}{r\cos \theta} \label{eq-5}
\end{equation}

From the equation we get two conclusions. First, $F_n \propto G$ 
when $0^\circ<\alpha<90^\circ$. 
Second, $d$ is irrelevant to $F_{n}$ (the force is the same as any grasping point).

The first conclusion further shows that a larger gravity
leads to a larger contact force between the robot gripper and the tool, and hence leads to
larger friction. That is, the gravity of the base frame contributes to increasing the 
friction between the robot gripper and the tool. The conclusion demonstrates 
another advantage of the configuration in Fig.\ref{parallel}(c)
over the one in Fig.\ref{parallel}(d).
The force relations of Fig.\ref{parallel}(d) is
\begin{equation}
F_n=-\frac{G\cos \alpha \tan \theta }{2}+\frac{2vF_{spring}}{r\cos \theta} \label{eq-6}
\end{equation}
where $F_{n} \propto -G$ when $0^\circ<\alpha<90^\circ$.
For this configuration, the gravity of the base frame reduces the friction and makes the hold unstable.
The configuration in Fig.\ref{parallel}(c) is preferable than the one in Fig.\ref{parallel}(d) 
when $0^\circ<\alpha<90^\circ$ (the tooltip faces downward).



\subsubsection{Grasping an object using the tool}
\begin{figure}[!htbp]
\begin{center}
\includegraphics[width=.9\columnwidth]{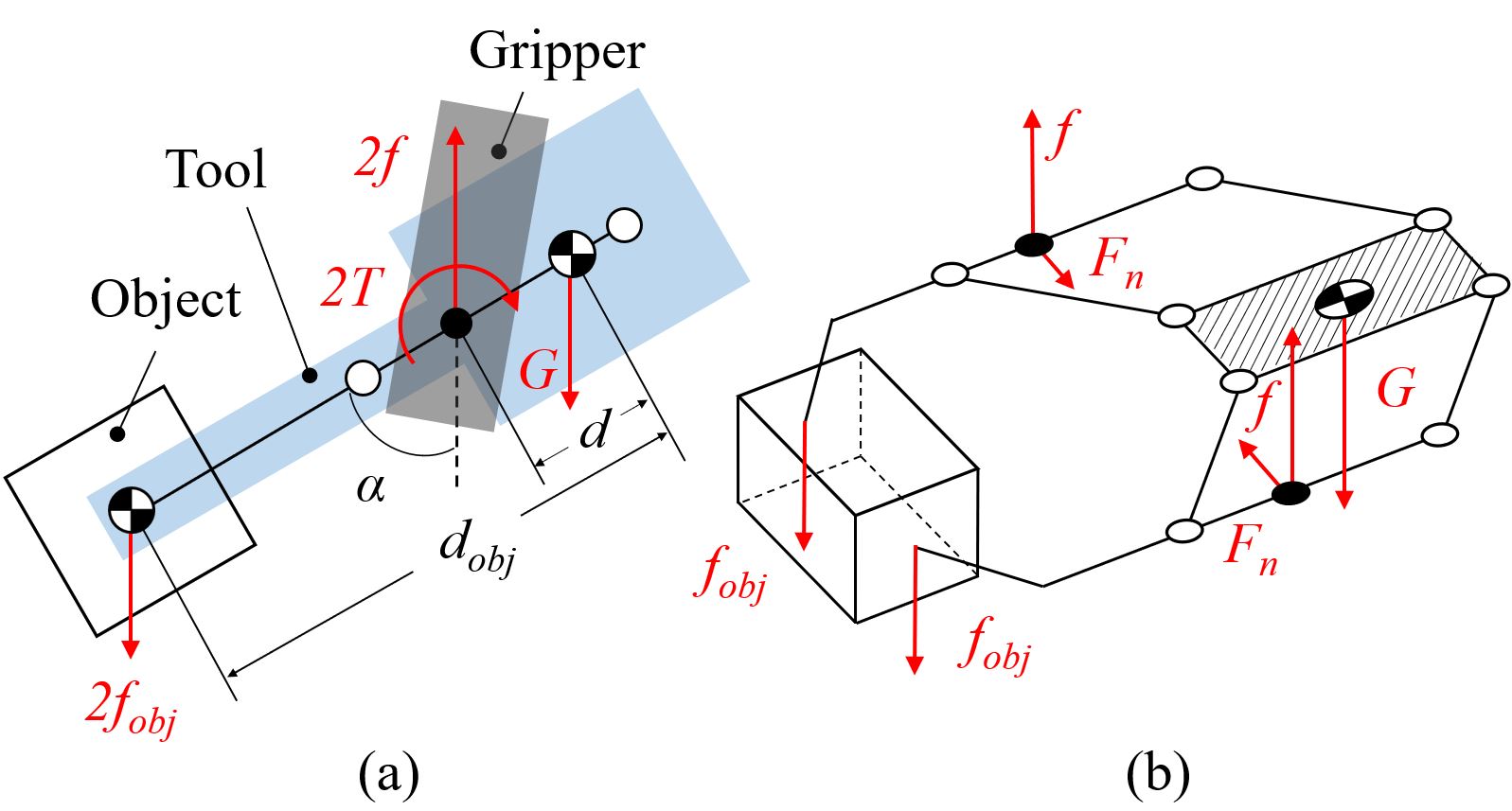}
\caption{The free body diagram when the tool is holding an object. 
$F_{n}$ is the force exerted by the robot gripper.}
\label{targetobj}
\end{center}
\end{figure}

We use the symbols shown in Fig.\ref{targetobj} to analyze the maximum weight of
objects that can be pick up by the tool.
To simplify the analysis, we assume the contact between the tooltips and the object
is co-linear with the $com$ of the object.

When the force and torque are balanced, we get:
\begin{equation}
2f-G-G_{obj} = 0 \label{eq-gobj-1}
\end{equation}
\begin{equation}
Gd_{com}\cos{\alpha}+2T-G_{obj}d_{obj}\sin{\alpha}=0 \label{eq-gobj-2}
\end{equation}

\begin{figure}[th!]
\begin{center}
\includegraphics[width=.9\columnwidth]{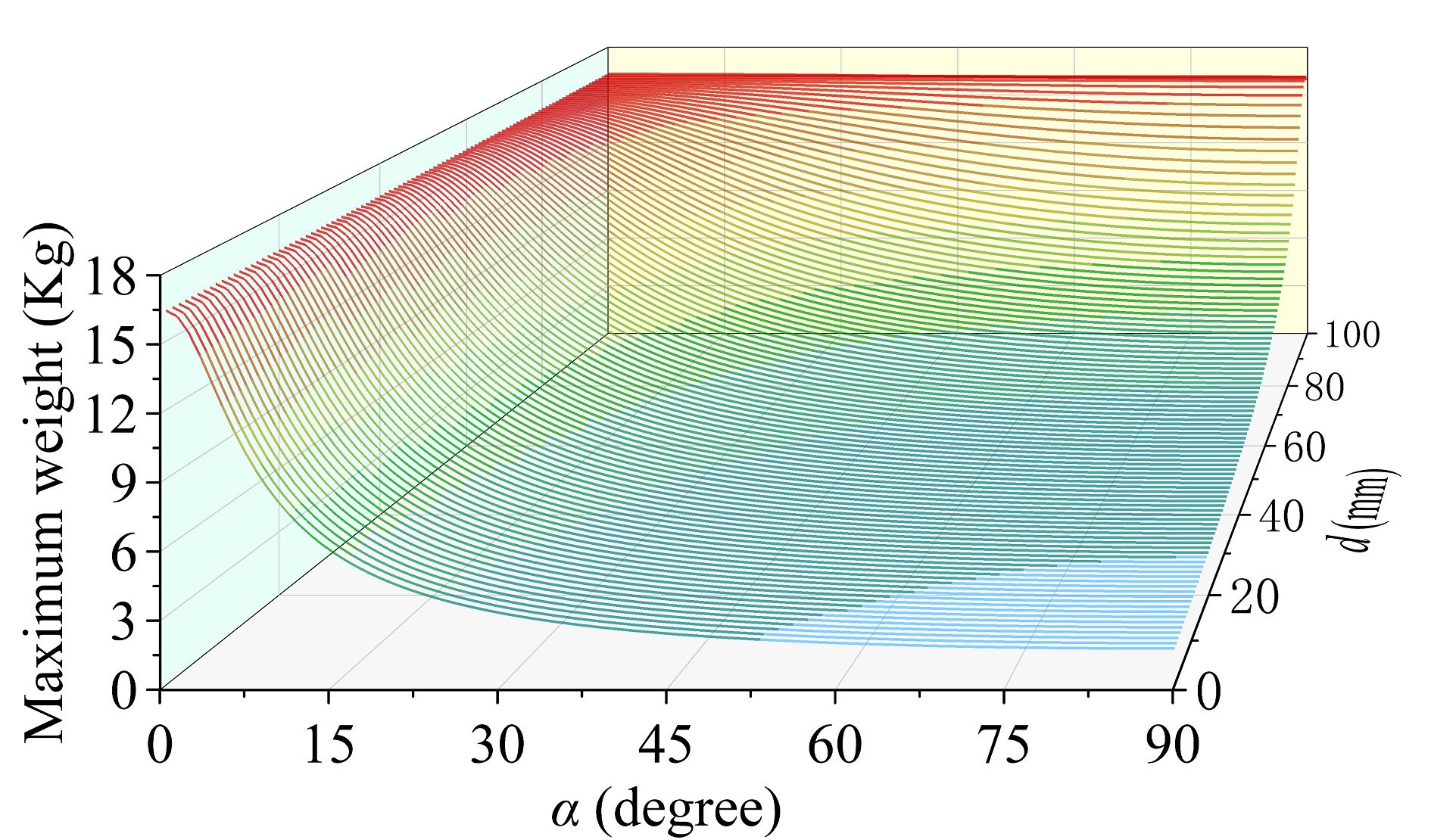}
\caption{The relation between the maximum weight of the object, the tool angle, and the tool grasping point.}
\label{force-analysis}
\end{center}
\end{figure}

The maximum weight of the object to be held can be computed using 
equations (\ref{eq-fa-1}), (\ref{eq-gobj-1}), and (\ref{eq-gobj-2}):
\begin{equation}
a=\frac{1+d_{obj}^{2}\sin{\alpha}^{2}\mu^{2}F_{n}^{2}}{4\max T^{2}} \label{a}
\end{equation}
\begin{equation}
b=\frac{\mu^{2}F_{n}^2(G-d_{obj}d\sin{\alpha}^{2})}{2\max T^{2}} \label{b}
\end{equation}
\begin{equation}
c=\frac{G^{2}(\max T^{2}+d^{2}\sin{\alpha}^{2}\mu^{2}F_{n}^{2})}{4\max T^{2}}-\mu^{2}F_{n}^{2}\label{c}
\end{equation}
\begin{equation}
Maximum~weight=\frac{-b+\sqrt{b^{2}-4ac}}{2a}\label{d}
\end{equation}
The maximum weight of the object is a function of the angle $\alpha$ and the distance $d$.
While the analytical form of the result is ugly, the 3D plot of the relationship is
shown in Fig.\ref{force-analysis} ($\max T$ is set to a constant value). 

\subsection{Size optimization}
\begin{figure}[!htbp]
\begin{center}
\includegraphics[width=.9\columnwidth]{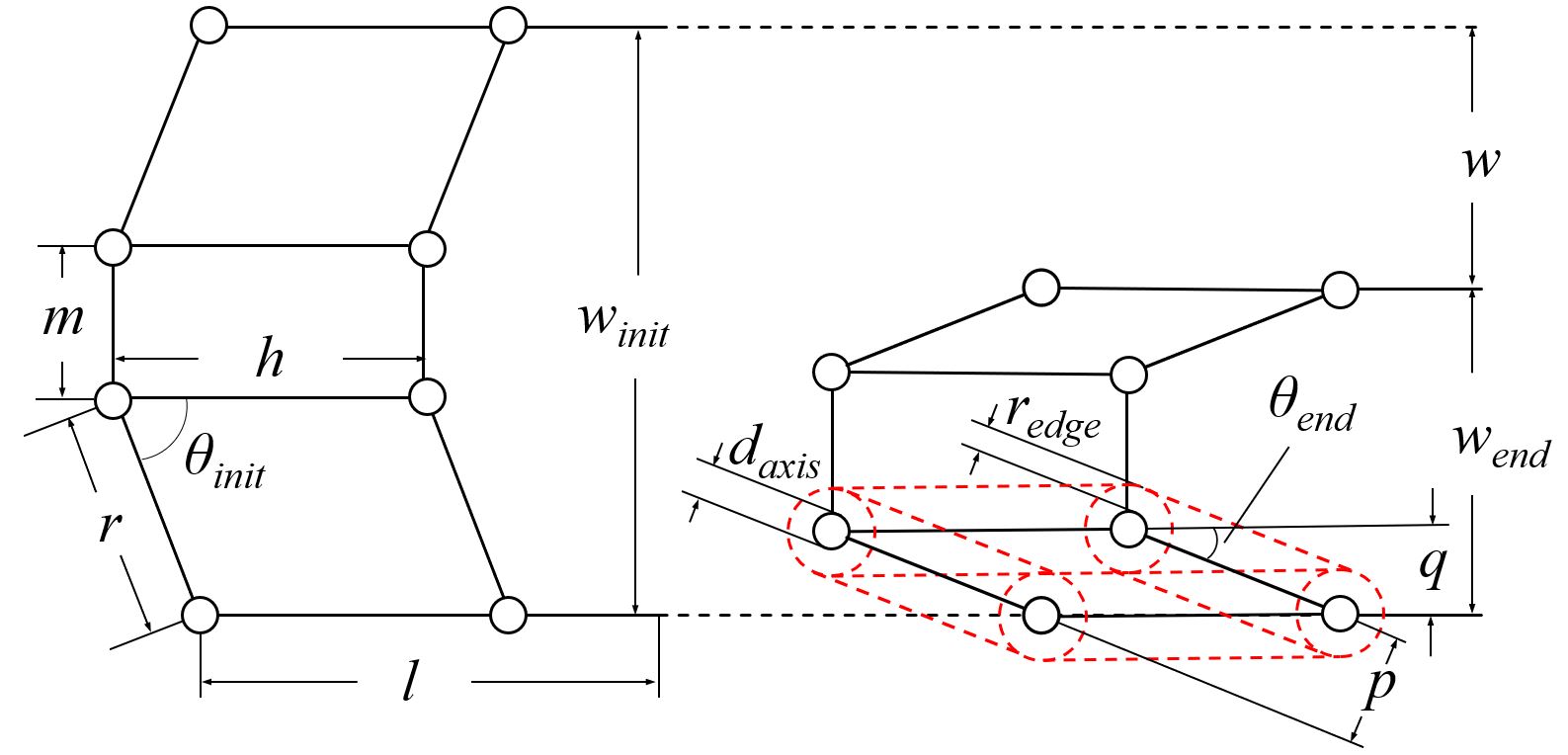}
\caption{The dimension parameters used to optimize the size of the tool.}
\label{dimension}
\end{center}
\end{figure}

In this section, we optimize the dimension of the tool.
We would like the tool to have a large stroke and compact size.
The dimension parameters shown in Fig.\ref{dimension} are used for optimization.

Equations (\ref{eq-1}) and (\ref{eq-2}) show the relationship between the width of the tool, the stroke of the tool,
and the rotational angles of the angular linkage.
\begin{equation}
w_{init} = m + 2r\sin\theta_{init} \label{eq-1}
\end{equation}
\begin{equation}
w=2r\sin\left ( \theta_{init} - \theta_{end} \right ) \label{eq-2}
\end{equation}
Given a fixed $w_{init}$, $w$ is expressed as:
\begin{equation}
w=\frac{l_{init}-m}{2sin\theta_{init}}\sin\left ( \theta_{init} - \theta_{end} \right ) \label{eq-3}
\end{equation}

Equation (\ref{eq-3}) shows that
increasing $\theta_{init}$ and decreasing $\theta_{end}$ and $m$ will enlarge the stroke $l$. 
When $\theta_{initi}$ is $90^{\circ}$, $\theta_{end}$ is $0^{\circ}$, and $m$ is $0$, $l$ reaches to its maximum. 
However, the relationship between $F_{n}$ and $\theta_{init}$ shown in (\ref{eq-5}) 
told that an overlarge $\theta_{initial}$ will significantly increase the requirements on the grasping force $F_n$
and make the tool hard to be compressed. If $\theta_{initi}$ reaches $90^{\circ}$, the tool 
can never be used. Also, if $m$ is 0, the base will disappear. In this case, since the links 
have a width in the real world, they are hard to be designed to move in the same plane 
unless increasing the complexity of the design and using an asymmetrical structure, 
which changes the position of the center of gravity and reduces the structural stability of 
the tool. In order to install the links appropriately, $m$ should meet:
\begin{equation}
m\geqslant d_{axis}+2r_{edge}
\end{equation}

For the same reason, $\theta_{end}$ cannot be $0^{\circ}$.
It is minimum can be calculated considering the radius of the joints and the width of the linkages.
The red dot lines in Fig.\ref{dimension}(b) show the situation when $\theta_{end}$ reaches its minimum.
In this case, the parallel linkage will touch the base frame. The minimum $\theta_{end}$ can be
computed using Equation (\ref{eq-size-3}). $p$ should meet equations (\ref{eq-size-4}) and (\ref{eq-size-5}). 
$h$ should meet equation (\ref{eq-size-6}). Otherwise, 
the link bars will overlap with each other.
\begin{equation}
\theta_{end}={\arcsin{\frac{q}{r}}}\label{eq-size-3}
\end{equation} 
\begin{equation}
q = d_{axis}+2r_{edge}\label{eq-size-4}
\end{equation} 
\begin{equation}
p\geqslant k\sin{\theta_{end}}\label{eq-size-5}
\end{equation} 
\begin{equation}
h\geqslant r\cos{\theta_{end}}+\tan{\theta_{end}(d_{axis}+2r_{edge})}\label{eq-size-6}
\end{equation} 


\subsection{Variation in tooltips}
\begin{figure}[!htbp]
\begin{center}
\includegraphics[width=.9\columnwidth]{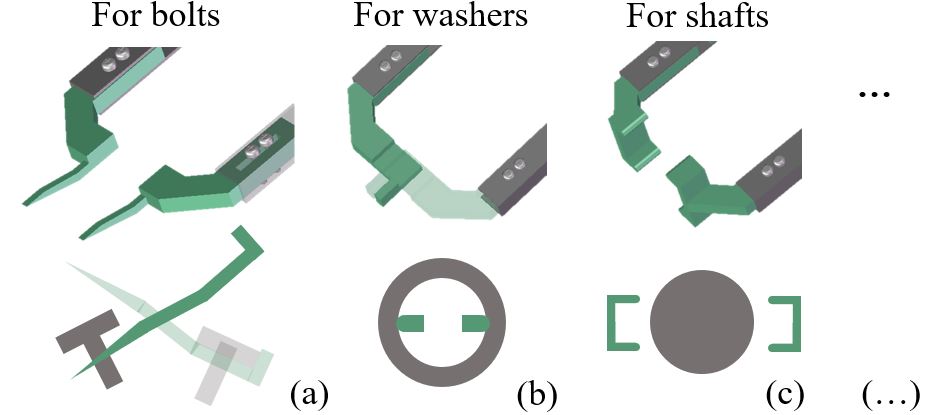}
\caption{Three exemplary special tooltips.}
\label{tooltips}
\end{center}
\end{figure}

In addition to the mechanical design,
we can make different tooltips for different tasks. Unlike the versatile gripper
which is designed to adapt to a wide range of tasks, each of the tooltip is specially designed for a single task. 
The design can thus be more compact and reliable. These different 
tooltips are based on the same mechanism so that a robot can manipulate them 
in the same way. Three examples of the special tooltips are shown in Fig.\ref{tooltips}. 

\section{Using the Tool}

When performing specific tasks, the tool is placed in an arbitrary pose in the work space. 
A robot identifies the tool using AR markers and grasps it using pre-planned grasp configurations.
To use the tool, the robot should be constrained to grasp the tool in specific poses (working poses).
When the pose of the tool makes it impossible or difficult to be picked up in a working pose,
regrasp planning \cite{wan2019preparatory} may be employed to adjust the grasp configuration.

\subsection{Working poses}

\begin{figure}[htbp]
\begin{center}
\includegraphics[width=.9\columnwidth]{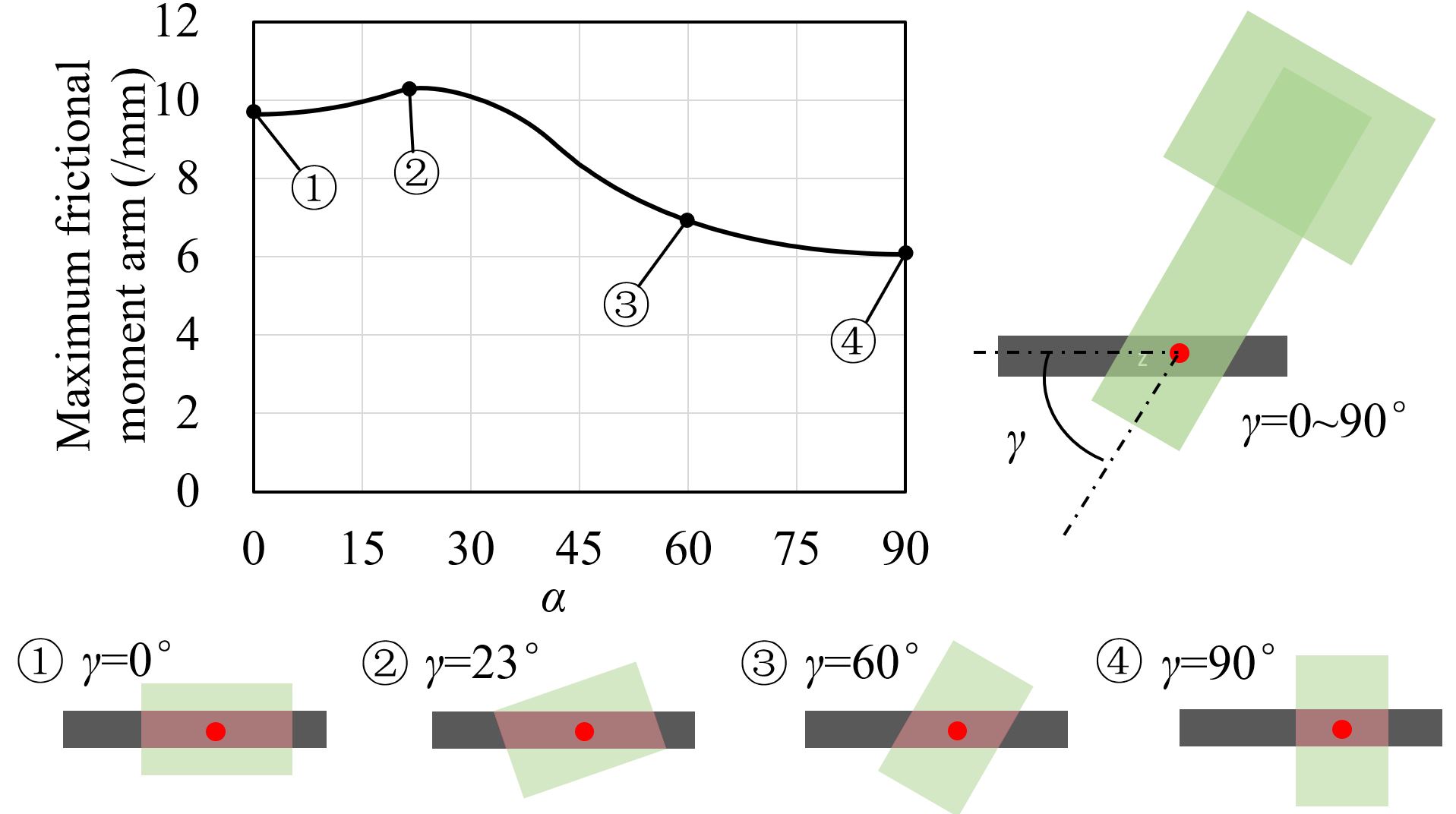}
\caption{The changes of maximum 
friction torque in the range of working poses.
The maximum frictional moment arm increases in the beginning, and reaches the peak at 23$^{\circ}$.}
\label{alphatorque}
\end{center}
\end{figure}

There are several pairs of parallel surfaces of the tools that can be stably grasped,
but the tool can be used only when the sides of the parallel linkages are used as the contact surfaces.
In addition, the angle $\alpha$ of the tool is expected to be within the range of $0^{\circ}\sim 90^{\circ}$
unless there are special requirements. The angle $\gamma$, which is defined as the
angle between the hand and the tool (Fig.\ref{alphatorque}), is also expected to be within
the range of $0^{\circ}\sim 90^{\circ}$.
The reason is two-fold. For one thing, $F_{n} \propto G$ when $0^\circ<\alpha<90^\circ$,
the tool is more stable. For the other, the gripper is facing backward and does not
obstruct the working space of the tool when $0^\circ<\gamma<90^\circ$.

Using the holding condition equation (\ref{eq-d-1}), the changes of maximum 
friction torque when $0^\circ<\gamma<90^\circ$ can be computed.
The result is shown in Fig.\ref{alphatorque} ($d$ is set to 0, $\alpha$ is set to 90$^\circ$.). 
The maximum friction torque increases in the beginning, 
and decreases after reaching the peak at $\gamma = 23^{\circ}$. 

\subsection{Reorienting to the working poses}

\begin{figure}[!htbp]
\begin{center}
\includegraphics[width=.9\columnwidth]{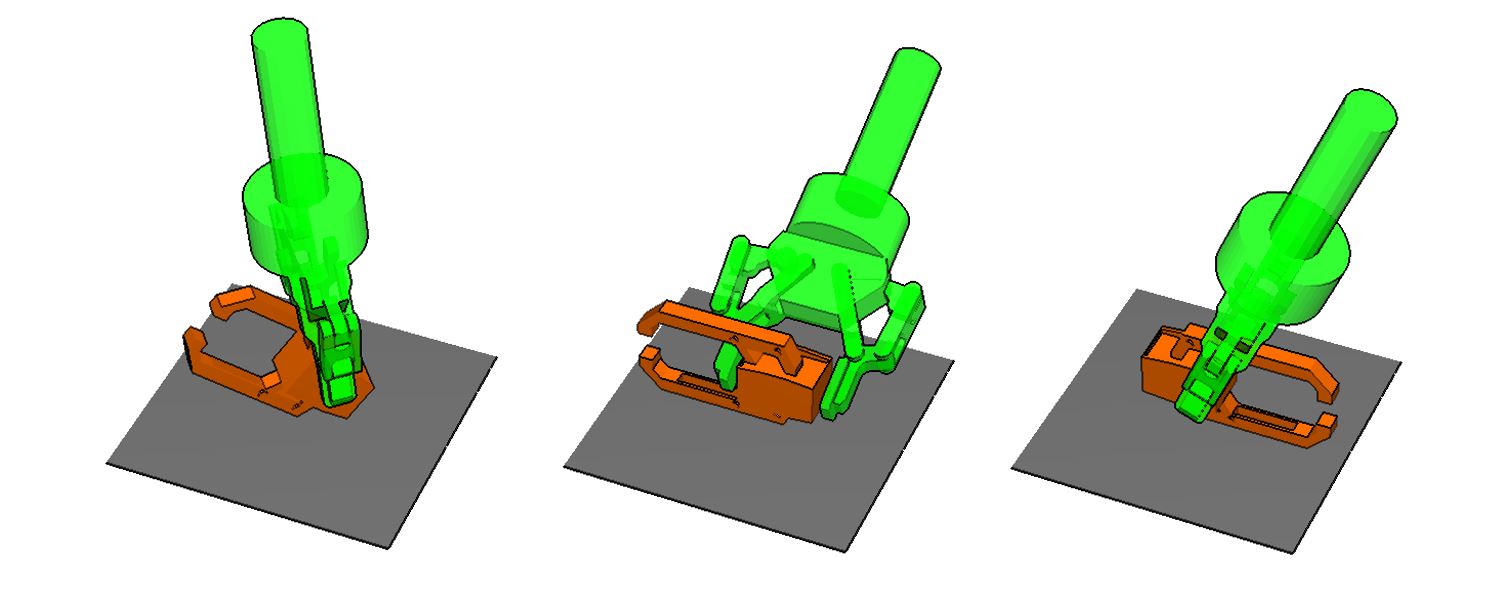}
\caption{Some poses of the tool that cannot be directly picked up into a working pose.}
\label{regrasp}
\end{center}
\end{figure}

The tool could be placed in an arbitrary pose in the work space.
The pose is not necessarily able to be picked up into a working pose.
For example, the poses shown in Fig.\ref{regrasp} can never be directly used since
the robot gripper can never grab the sides of the parallel linkages.
In that case, reorienting planners are needed to help the robot adjust the poses.
The authors had been working on reorienting planners for several years. This paper uses
one of our recent results (regrasp using two arms) \cite{} to plan reorienting tools.

\section{Experiments}

\begin{figure}[!htbp]
\begin{center}
\includegraphics[width=.8\columnwidth]{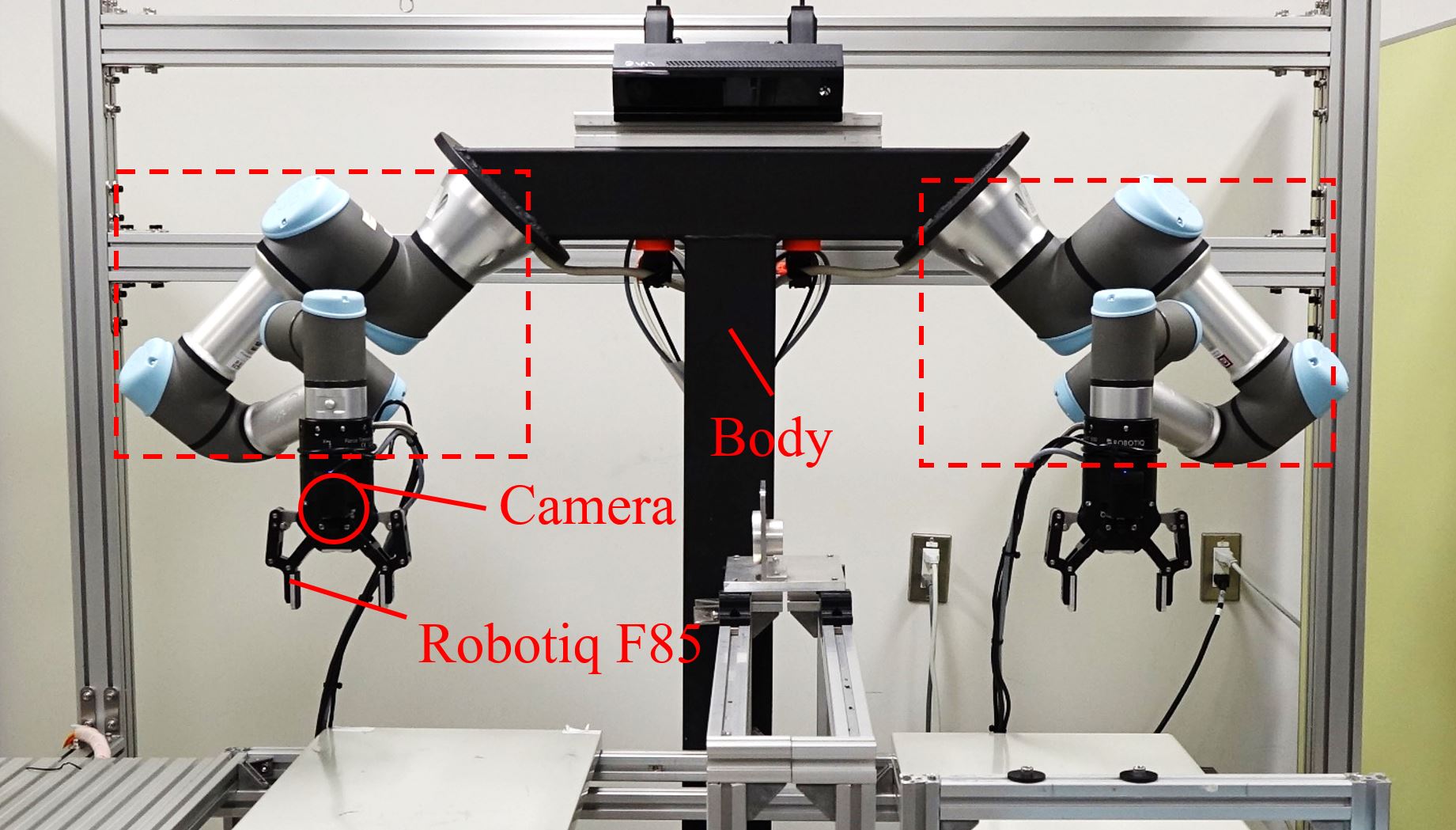}
\caption{The experiment system. It has a dual-arm UR3 robot with Robotiq F85 grippers 
installed to both of them. Cameras are also mounted on both grippers.
}
\label{robot}
\end{center}
\end{figure}

Fig.\ref{robot} shows our experiment system. We used a dual-arm UR3 
robot to conduct the experiments. Two UR3 robots were mounted symmetrically 
with $45^{\circ}$ to the body frame. Robotiq F-85, a two-finger 
parallel-jaw gripper, was employed for both arms. For visual detection, a camera, ELP-USBFHD06H-L36 skewless HD, 
was mounted to one side of Robotiq F-85.

\subsection{Maximum weight of an object}

\begin{figure}[!htbp]
\begin{center}
\includegraphics[width=.8\columnwidth]{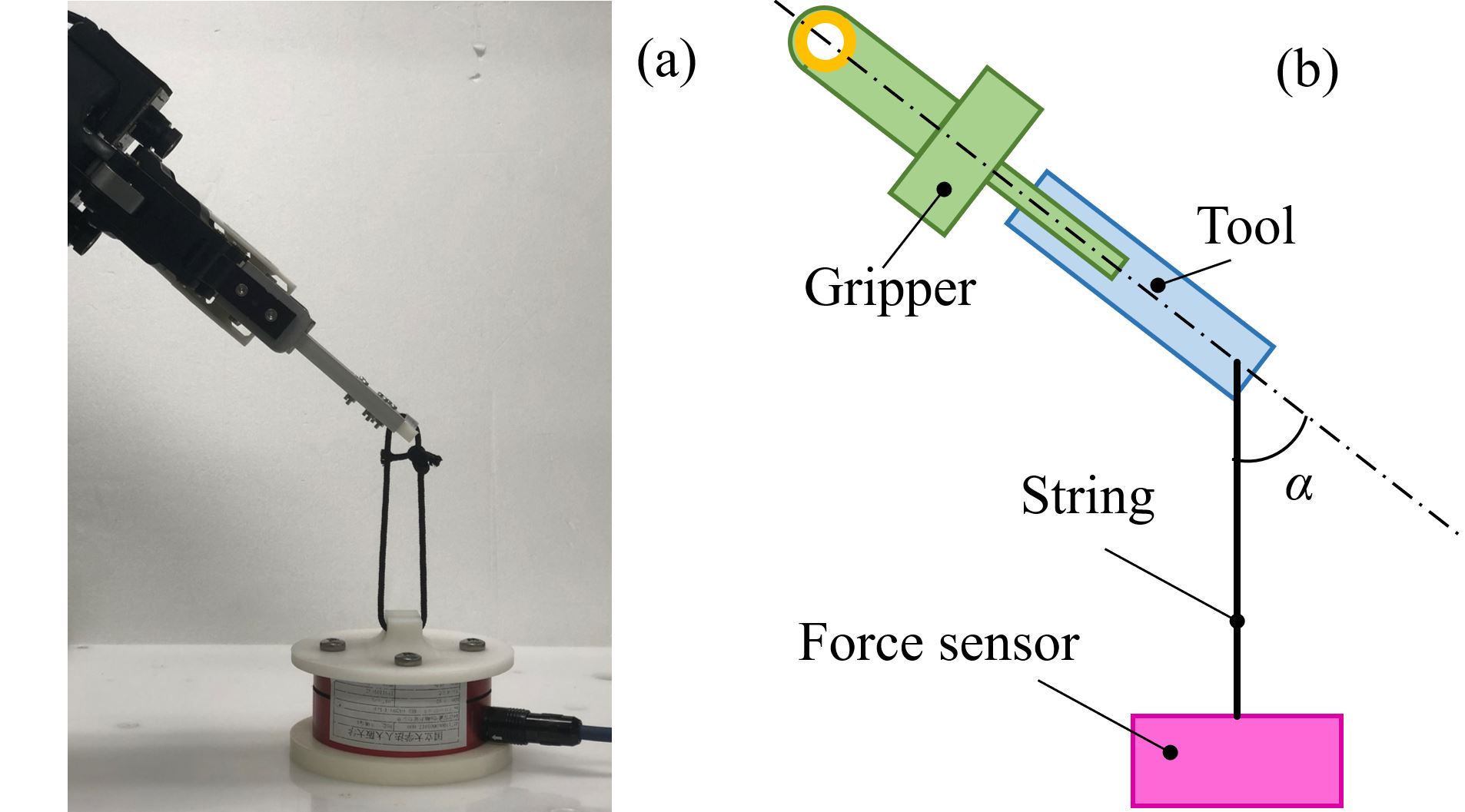}
\caption{The set up to measure the maximum weights of objects that
can be picked by the tool. The peak force measured by the force sensor before the tool moves is
recorded as the maximum weight of objects that can be picked by the tool.
}
\label{ft}
\end{center}
\end{figure}

First, we perform experiments to test the maximum weights of objects that
can be picked by the tool. DynPick force sensor is used for measurement.
The setting is shown in Fig.\ref{ft}. The sensor is fixed to a table.
A string is used to connect the sensor and the tooltips. 
The robot gripper is used to hold the tool and drag the string up vertically until the tool is moved.
The peak force measured by the force sensor before the tool moves is
recorded as the maximum weight of objects that can be picked by the tool.
The test is repeated with the $\alpha$ angle changing from 0$^\circ$ to 75$^\circ$.
To simplify the experiment and 
improve precision, the $\gamma$ angle is set to $0^{\circ}$. 

\begin{figure}[h]
\begin{center}
\includegraphics[width=.85\columnwidth]{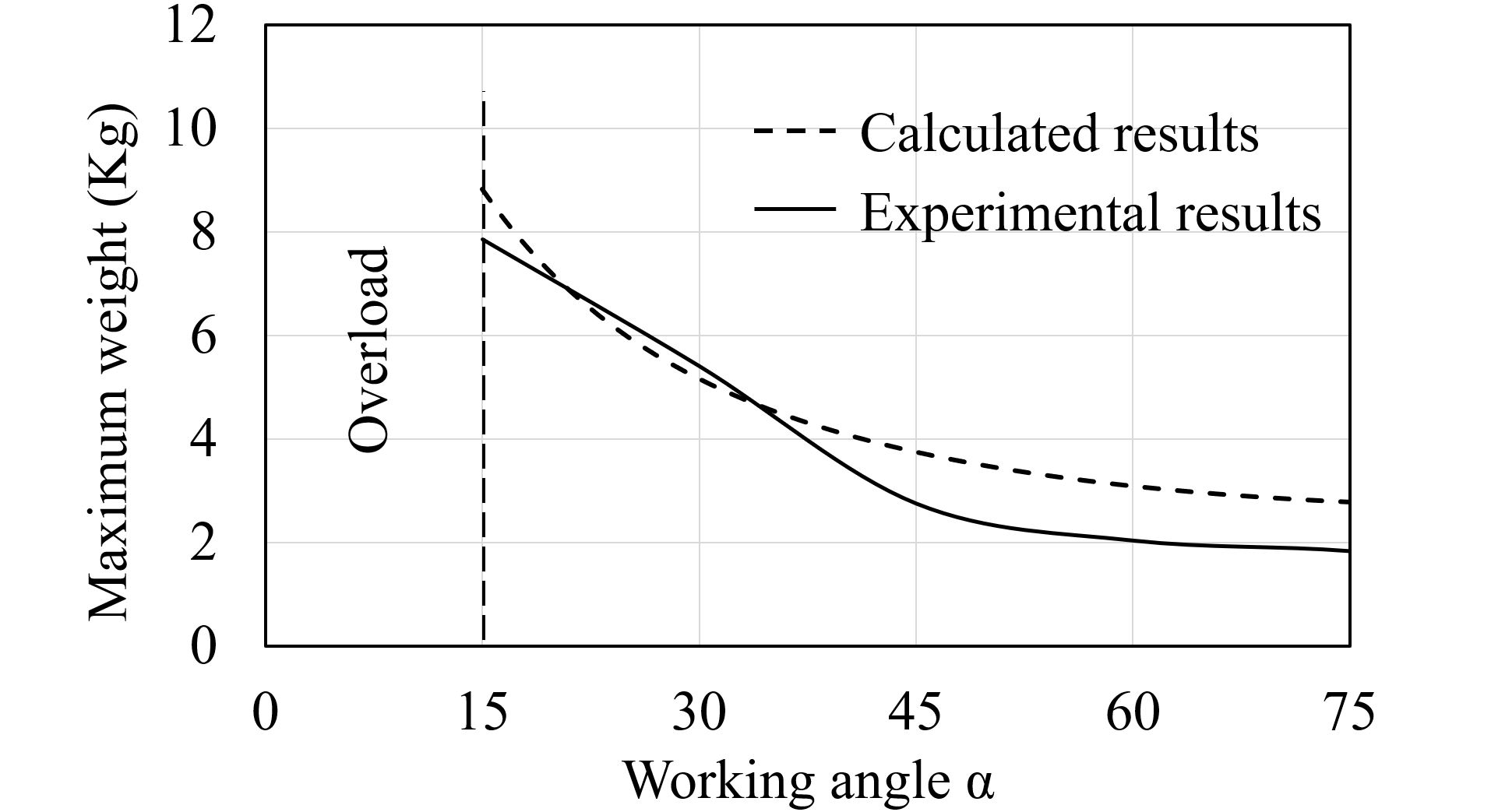}
\caption{The maximum weight that the tool can pick at different $\alpha$.}
\label{force1}
\end{center}
\end{figure}

The result is shown in Fig.\ref{force1}. The solid curve shows the data
measured by the experiments. The dash curve is the computed value using equation (\ref{a}-\ref{d}).
The measured results are nearly the same as the theoretical analysis. The proposed design
could pick up an object of 8 $kg$ when the angle between the tooltip and the gravity direction is 15$^\circ$.
 
\subsection{Using the tool to pick various objects}
\begin{figure}[th!]
\begin{center}
\includegraphics[width=.9\columnwidth]{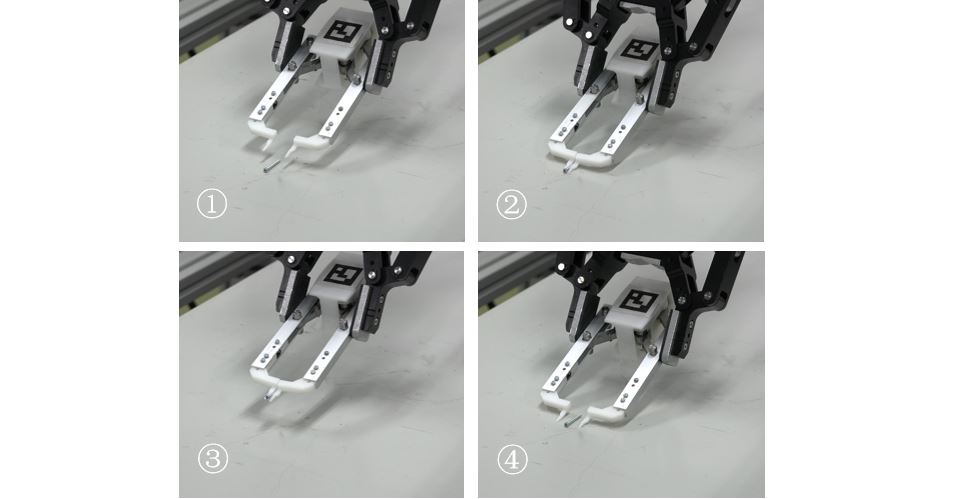}
\caption{Using the tool to pick up a screw.}
\label{demo-3}
\end{center}
\end{figure}

\begin{figure}[th!]
\begin{center}
\includegraphics[width=.9\columnwidth]{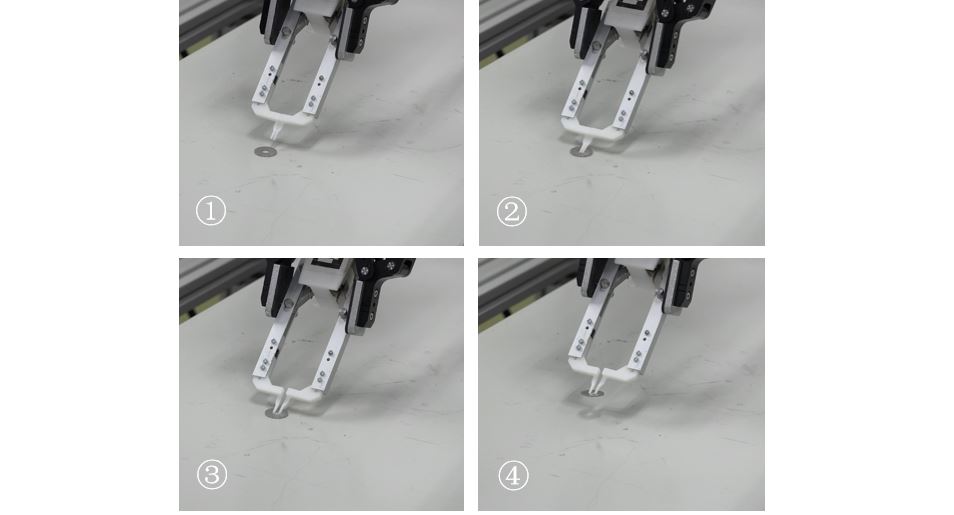}
\caption{Using the tool to grasp and pick up a washer.}
\label{demo-4}
\end{center}
\end{figure}

Second, to test the performance of the tool, we conducted experiments to pick up 
a screw, a washer, and a can. These objects are difficult to be handled 
by only the Robotiq F85 gripper. The screw and washer are too small to be 
grasped by the gripper, successfully handling them would demonstrate the feasibility of 
the tool for picking up small mechanical parts. When the screw is
on the table, there is limited grasping space and requires suitable gripper 
shape and precise position control. Fig.\ref{demo-3} shows the experimental results
of handling a screw. In this case, a thin tooltip was used \cite{nie2019hand}. The 
compliant units at the parallel linkage allowed contacting 
the table completely to achieve an available grasp. 
Compared to the screw, the washer is even more difficult to grasp due to its
thin thickness and circular outside. The tooltip in this case was designed to stretch
against the inner circle of the washer (radius: $1.5 mm$). the tool was compressed fully to insert 
the tooltips into the inner circle. After insertion, 
the gripper opened a bit to hold and pick up the washer, as is shown in Fig.\ref{demo-4}.


\begin{figure}[th!]
\begin{center}
\includegraphics[width=.95\columnwidth]{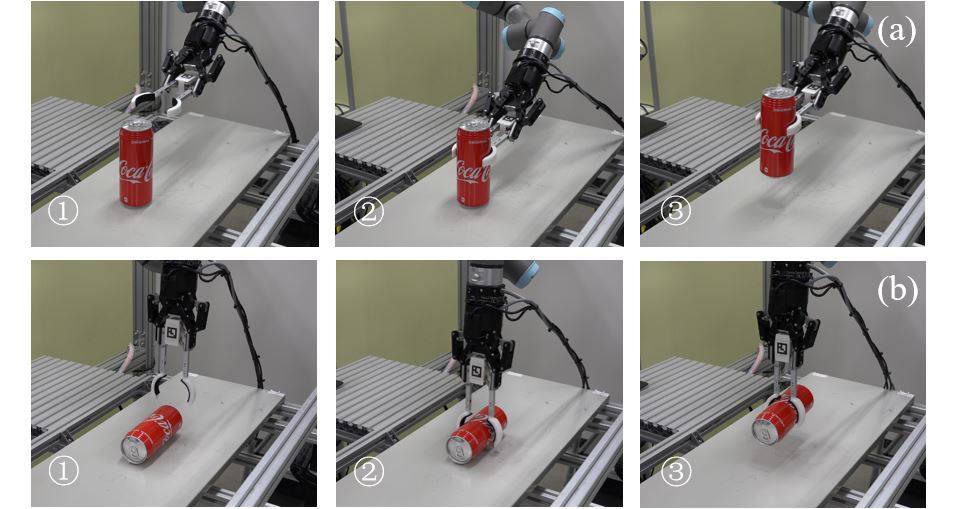}
\caption{(a) Using the tool to grasp and pick up a can from an upright pose.
(b)Using the tool to grasp and pick up a can from an lateral pose.}
\label{demo-5}
\end{center}
\end{figure}

The third object, namely the can, is easy to deform. Grippers without feedback 
control are likely to crash the can. With a properly designed tooltip, the
proposed tool has the merit to safely pick up a can. 
Fig.\ref{demo-5} shows two examples. A can has two stable placement poses, and 
both of them can be handled using the tool.

\subsection{Automatic recognition and planning}
\begin{figure*}[!htbp]
\begin{center}
\includegraphics[width=.99\textwidth]{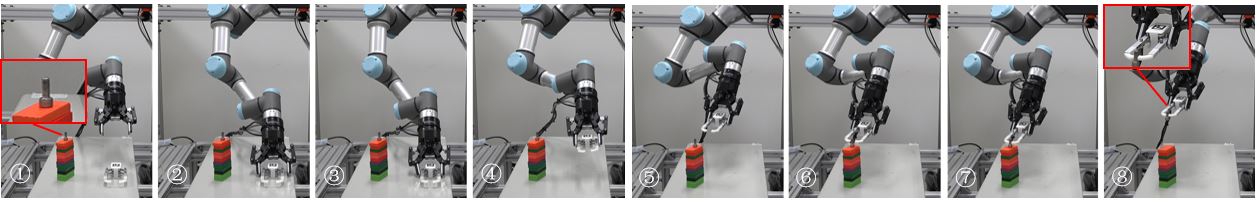}
\caption{Automatic recognition and planning 1. In this case, the 
tool can be manipulated into a working pose to pick up the bolt without any adjustment.}
\label{bolt}
\end{center}
\end{figure*}

\begin{figure*}[!htbp]
\begin{center}
\includegraphics[width=.99\textwidth]{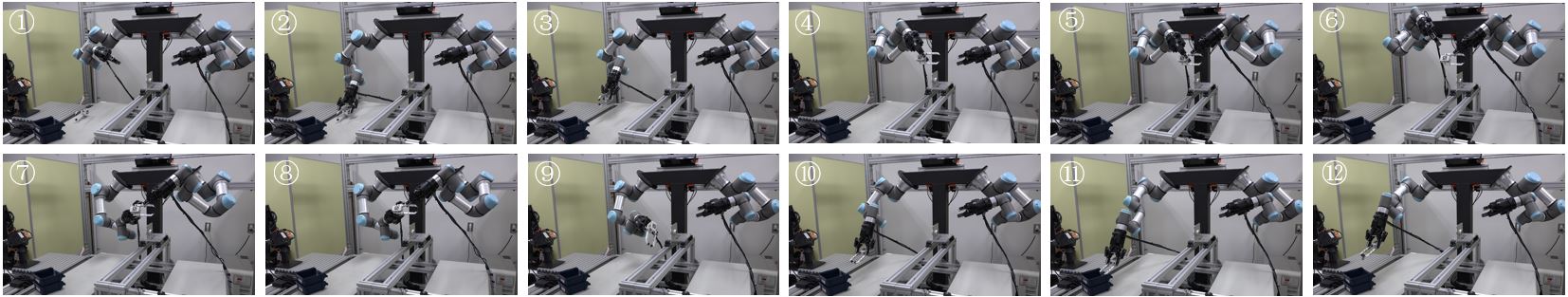}
\caption{Automatic recognition and planning 2. In this case, the 
tool cannot be manipulated into a working pose directly. The robot used a handover
to adjust the pose of the tool.}
\label{bolt2}
\end{center}
\end{figure*}

Third, we used the tool to pick up a bolt with automatic planning. 
Fig.\ref{bolt} is the result of the first test. In this case,
the tool can be picked up into a working pose without any adjustment.
The robot recognized the tool, picked it up, planned a motion to use the tool to pick up the bolt.
Fig.\ref{bolt2} is the result of the second test. In this case, the
tool cannot be picked up into a working pose directly. Regrasp planning is
used to adjust the pose of the tool.
The robot recognized the tool, planned a regrasp motion to
adjust the pose of the tool (a handover was used to adjust the pose of the tool into a working pose),
and planned a motion to use the tool to pick up the bolt. Together with the 
automatic recognition and planning the tool can be used flexibly without
requirements on power supply, vacuum supply, or delicate mechanism and control. 

\section{Conclusions and Future Work}

This paper presented a mechanical tool designed for robots.
The idea is inspired by the tools designed for human hands.
The tool is purely mechanical. It is free of power supply, vacuum supply, and delicate mechanism and control.
It is actuated by the gripper force from the robot gripper.
The tool may have various tooltips designed for specific tasks 
to extend the function of robot grippers. 
The quantitative analysis and real-world experiments demonstrate the feasibility 
of the tool and show that it is able to be held and manipulated by a robot gripper and 
to realize various picking tasks. Our future work includes diversifying 
the tooltips to develop wider applications as well as optimizing the planning algorithms. 

\bibliographystyle{IEEEtran}
\bibliography{huiros2019}

\end{document}